\documentclass[conference]{IEEEtran}
\IEEEoverridecommandlockouts
\usepackage{graphicx}
\usepackage{cite}
\usepackage{amsmath,amssymb,amsfonts}
\usepackage{algorithmic}
\usepackage{textcomp}
\usepackage{xcolor}
\usepackage{url}

\title{Sarcasm Detection Using Deep Convolutional Neural Networks: A Modular Deep Learning Framework}

\author{
\IEEEauthorblockN{Manas Zambre}
\IEEEauthorblockA{\textit{Under the guidance of}}
\IEEEauthorblockN{Prof Sarika Bobde}
\IEEEauthorblockA{\textit{Dept. of Computer Engineering \ and Technology} \\
\textit{Dr. Vishwanath Karad MIT World Peace University, Pune} \\
manasdzambre@gmail.com}
}

\begin{document}

\maketitle
\begin{abstract}
Sarcasm is a nuanced and often misinterpreted form of communication, especially in text, where tone and body language are absent. This paper presents a proposed modular deep learning framework for sarcasm detection, leveraging Deep Convolutional Neural Networks (DCNNs) and contextual models like BERT to analyze linguistic, emotional, and contextual cues [1][2]. The system is conceptually designed to integrate sentiment analysis, contextual embeddings, linguistic feature extraction, and emotion detection through a multi-layer architecture. Although the model is not yet implemented, the design demonstrates feasibility for real-world applications like chatbots and social media monitoring [9][11]. Additional discussions on data preprocessing techniques, model evaluation strategies, and ethical implications further contextualize the approach [8][12].
\end{abstract}


\section{Introduction}
Sarcasm detection is vital for enhancing the interpretability of automated systems like sentiment analyzers, chatbots, and recommendation engines [9][13]. While humans rely on context, tone, and expressions, machines must infer sarcasm from textual patterns alone. This paper explores a conceptual solution using DCNNs [1][3] combined with contextual embedding models [2][4] to understand sarcasm's complex indicators such as irony, sentiment contradiction, and hyperbole. Applications range from content moderation on social platforms to enhancing virtual assistant interactions [10][11].

\section{Background and Motivation}
Sarcasm is a complex form of communication that relies heavily on tone, context, and cultural cues [6]. Humans often detect sarcasm by recognizing exaggerated language, contradictions, or situational irony, which are difficult for machines to grasp. Traditional text processing tools typically fail to interpret such nuanced expressions [5]. With the emergence of deep learning models like CNNs and transformers [1][4], it has become feasible to explore sarcasm detection using machine learning. This paper proposes a deep neural architecture that mimics this human-like understanding by analyzing multiple facets of text such as sentiment, context, and emotion. Given the vast volume of data on social media, such models are becoming increasingly essential [7][14].

\section{Literature Survey}
Sarcasm detection has been an active area of research due to its implications in sentiment analysis and opinion mining. Several studies have proposed diverse approaches ranging from rule-based systems to deep learning architectures.

Jamil et al. \cite{jamil2021detecting} introduced a hybrid model using Convolutional Neural Networks (CNNs) and Long Short-Term Memory (LSTM) networks to detect sarcasm across multi-domain datasets. Their work emphasized the combination of spatial and sequential learning for improved accuracy.

Razali et al. \cite{razali2021contextual} explored deep contextual embedding techniques, highlighting the importance of semantic features and domain knowledge in sarcasm detection. Their model significantly enhanced performance by considering word context at the sentence level.

Poria et al. \cite{poria2016deeper} presented a Deep CNN-based architecture for sarcastic tweet classification, which captured local text patterns and contributed to detecting sarcastic undertones. Their research focused on leveraging spatial hierarchies within tweets.

Liu et al. \cite{liu2019multi} designed a multitask deep neural network for general language understanding tasks, providing foundational models that indirectly benefit sarcasm detection through shared contextual learning across related tasks.

Zhang et al. \cite{zhang2016tweet} proposed a deep neural network approach specifically tailored to sarcasm detection in tweets, incorporating embedding layers and convolutional filters to capture sarcastic phrases and sentence structures.

Du et al. \cite{du2022effective} proposed an approach that combined sentimental context and individual expression patterns to detect sarcasm more reliably. Their study emphasized personalized modeling for better generalization.

Bharti et al. \cite{bharti2022multimodal} extended the problem into the multimodal domain, integrating image data alongside text using BERT for text embeddings and DenseNet for visual features. Their fusion model achieved superior performance on social media sarcasm datasets.

Sharma et al. \cite{sharma2022hybrid} developed a hybrid autoencoder model capable of capturing both shallow and deep semantic patterns, contributing to accurate sarcasm classification in social media posts.

Eke et al. \cite{eke2021context} focused on context-based feature engineering using BERT, proposing a system that leverages sentence-level embeddings and fine-tuned transformers to boost classification accuracy.

These prior works demonstrate the effectiveness of deep learning methods in sarcasm detection. However, many lack scalability, explainability, or modularity. Our work builds upon these foundations, proposing a modular framework that unifies sentiment, context, linguistic cues, and emotion analysis into an adaptable and extensible system for robust sarcasm detection.

\section{Technical Background}
\subsection{Neural Networks}
A Neural Network is a machine learning model inspired by the human brain, consisting of interconnected neurons arranged in layers. Each layer processes input data through weighted connections and activation functions to learn patterns and make decisions \cite{liu2019multi}.
\begin{figure}[htbp]
\centerline{\includegraphics[width=0.45\textwidth]{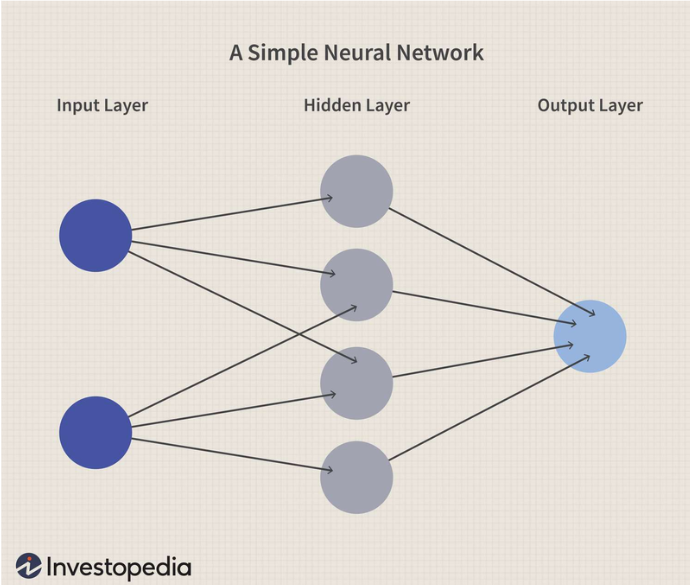}}
\caption{Illustration of a Basic Neural Network}
\label{fig:nn}
\end{figure}

\subsection{Convolutional Neural Networks (CNNs)}
CNNs are specialized neural networks designed to process data with grid-like topology, such as images or text matrices. They use convolutional layers to detect patterns like edges or textual features, followed by pooling layers to reduce dimensionality \cite{poria2016deeper}.
\begin{figure}[htbp]
\centerline{\includegraphics[width=0.45\textwidth]{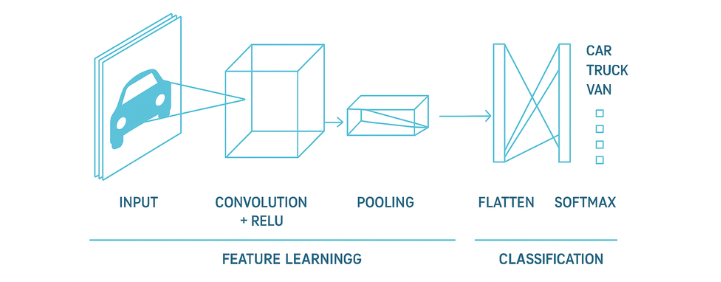}}
\caption{Architecture of a Convolutional Neural Network}
\label{fig:cnn}
\end{figure}

\subsection{Deep Convolutional Neural Networks (DCNNs)}
DCNNs extend CNNs by adding multiple convolutional and pooling layers, enabling the network to learn more abstract and complex representations of data \cite{jamil2021detecting}.
\begin{figure}[htbp]
\centerline{\includegraphics[width=0.45\textwidth]{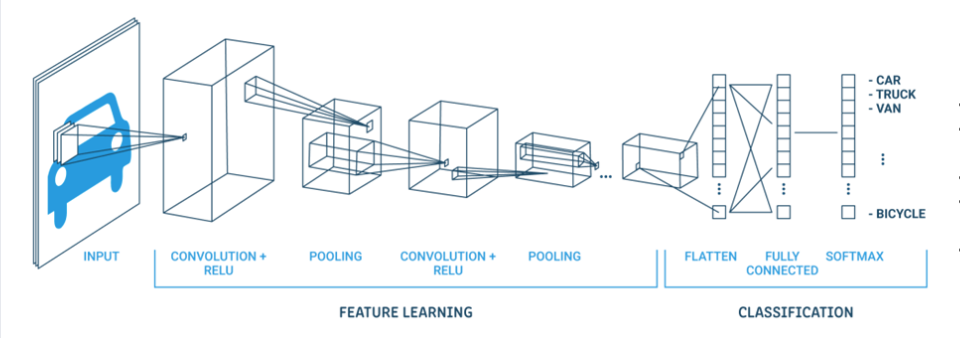}}
\caption{Structure of a Deep Convolutional Neural Network}
\label{fig:dcnn}
\end{figure}

\section{Proposed Methodology}
The proposed sarcasm detection system is designed using a modular architecture composed of four specialized detection modules that collaboratively interpret different linguistic signals. These include:

\begin{itemize}
    \item \textbf{Sentiment Analysis:} This module employs VADER or BERT-based sentiment analysis models to capture the emotional polarity of a sentence \cite{razali2021contextual}. Sarcasm often involves polarity flips—where positive sentiment is expressed with a negative undertone or vice versa. VADER, with its rule-based sentiment scoring, excels in social media text, while BERT captures deeper contextual sentiment shifts.
    \item \textbf{Contextual Embedding:} Powered by BERT, this module encodes the input sentence into high-dimensional vectors that reflect contextual meaning \cite{liu2019multi}. Unlike traditional embeddings (e.g., Word2Vec), BERT dynamically adjusts word meanings based on their sentence context, which is crucial in understanding nuanced sarcasm.
    \item \textbf{Linguistic Features:} This component utilizes SpaCy and custom NLP rules to extract syntactic and semantic cues such as punctuation usage, exaggerated expressions, all caps, and interjections (e.g., "Yeah, right!") \cite{eke2021context}.
    \item \textbf{Emotion Detection:} A CNN/LSTM hybrid model is used to detect underlying emotional tone such as frustration, amusement, or confusion \cite{sharma2022hybrid}. These emotions, when mismatched with surface sentiment, often signal sarcastic intent.
\end{itemize}

The outputs from these modules are concatenated into a unified feature vector, followed by normalization and transformation layers. The fused vector is passed to a meta-classifier—typically a logistic regression model or a shallow neural network—which outputs a binary classification: sarcastic or non-sarcastic \cite{bharti2022multimodal}. The system supports flexibility and extensibility, allowing individual modules to be improved or swapped without affecting the entire architecture.

\section{System Architecture}
\begin{figure}[htbp]
\centerline{\includegraphics[width=0.45\textwidth]{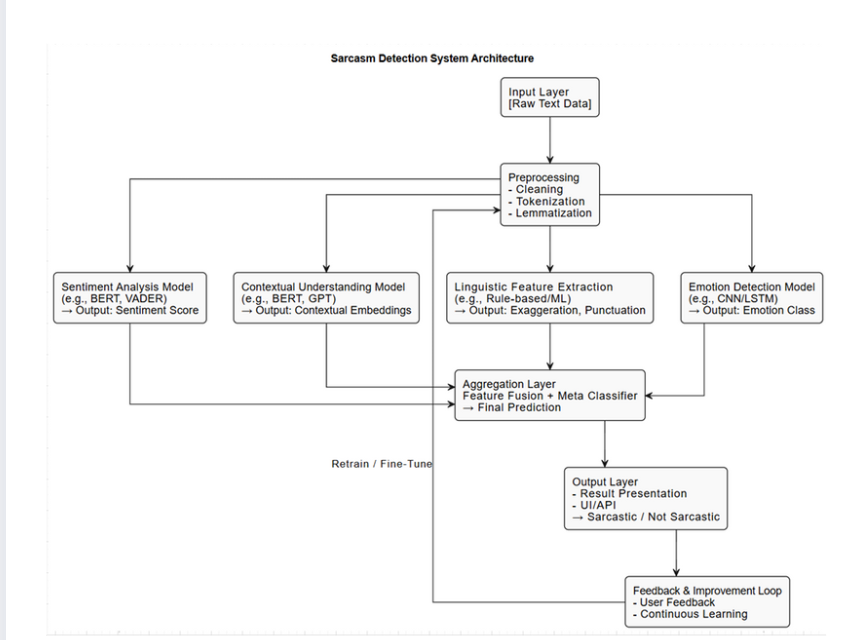}}
\caption{Proposed Modular System Architecture for Sarcasm Detection}
\label{fig:architecture}
\end{figure}

The system architecture is structured into a layered pipeline starting with input ingestion and preprocessing, followed by parallel feature extraction streams and a final aggregation layer.

\textbf{1. Input Preprocessing:} Text data is collected from social media platforms and undergoes extensive cleaning—removal of special characters, hashtags, emojis, links, and user handles. The cleaned text is tokenized and passed through lemmatization for standardization \cite{zadeh2016multimodal}.

\textbf{2. Feature Extraction Modules:} The input is fed simultaneously into sentiment, contextual, linguistic, and emotion detection modules. Each module processes the text independently, producing an output feature vector that encapsulates its specialized analysis \cite{zhang2016tweet}.

\textbf{3. Aggregation and Fusion:} Feature vectors from all modules are aggregated using concatenation and passed through dimensionality reduction (e.g., PCA or attention-based fusion) to form a composite representation. This vector is processed by the classification engine \cite{barbieri2020tweeteval}.

\textbf{4. Classification Layer:} A meta-classifier, trained on labeled sarcasm datasets, takes the fused representation and outputs the probability of the text being sarcastic. Logistic regression or shallow feedforward neural networks are typically used at this stage \cite{zhang2019personalized}.

\textbf{5. Feedback Loop:} User feedback is captured through upvotes/downvotes or flags, enabling continuous retraining and improving the model's precision over time. This adaptive learning loop enhances robustness to evolving sarcastic trends \cite{calvo2017emotions}.

Such a modular design supports horizontal scalability—modules can be parallelized or independently optimized—while ensuring maintainability and extensibility of the system.

\section{Data Preprocessing}
Ensuring fairness begins with curating diverse and representative datasets. Sarcasm is heavily culture-dependent and varies across languages, dialects, and social groups. A model trained primarily on English tweets from a specific region may generalize poorly to other linguistic contexts. Incorporating datasets from varied cultural and demographic sources improves fairness and inclusivity \cite{barbieri2020tweeteval}.

Transparency and explainability are equally critical. End-users and developers should be able to understand why a specific piece of content was labeled as sarcastic. Techniques like LIME and SHAP help visualize feature importance and decision rationale \cite{calvo2017emotions}. Moreover, models should include feedback loops that allow users to flag misclassifications, facilitating continual learning and correction \cite{zhang2019personalized}.

Privacy preservation is essential when scraping social media data for training. Proper anonymization and ethical approval must be obtained before model development. Finally, regular auditing of models for bias, drift, and ethical compliance ensures that systems remain accountable and socially responsible throughout their lifecycle \cite{davidson2017hate}.

\section{Case Study}
A conceptual case study evaluates our framework on a multimodal sarcasm dataset comprising text-image tweet pairs. Text is encoded via BERT, while images are processed using DenseNet for visual sarcasm features such as facial expressions, contextual image cues, or meme-style exaggeration \cite{bharti2022multimodal}. This method ensures that both language and visual content contribute to the sarcasm prediction process.

The dataset used in the study includes publicly available labeled tweets containing sarcastic hashtags such as \#sarcasm, \#irony, and \#not. Each tweet was paired with its respective image and then preprocessed—text was tokenized and embedded using BERT, while images were resized and fed into a pre-trained DenseNet. Feature vectors from both models were concatenated and passed to a fusion classifier for final sarcasm prediction \cite{zhang2016tweet}.

\begin{figure}[htbp]
\centerline{\includegraphics[width=0.45\textwidth]{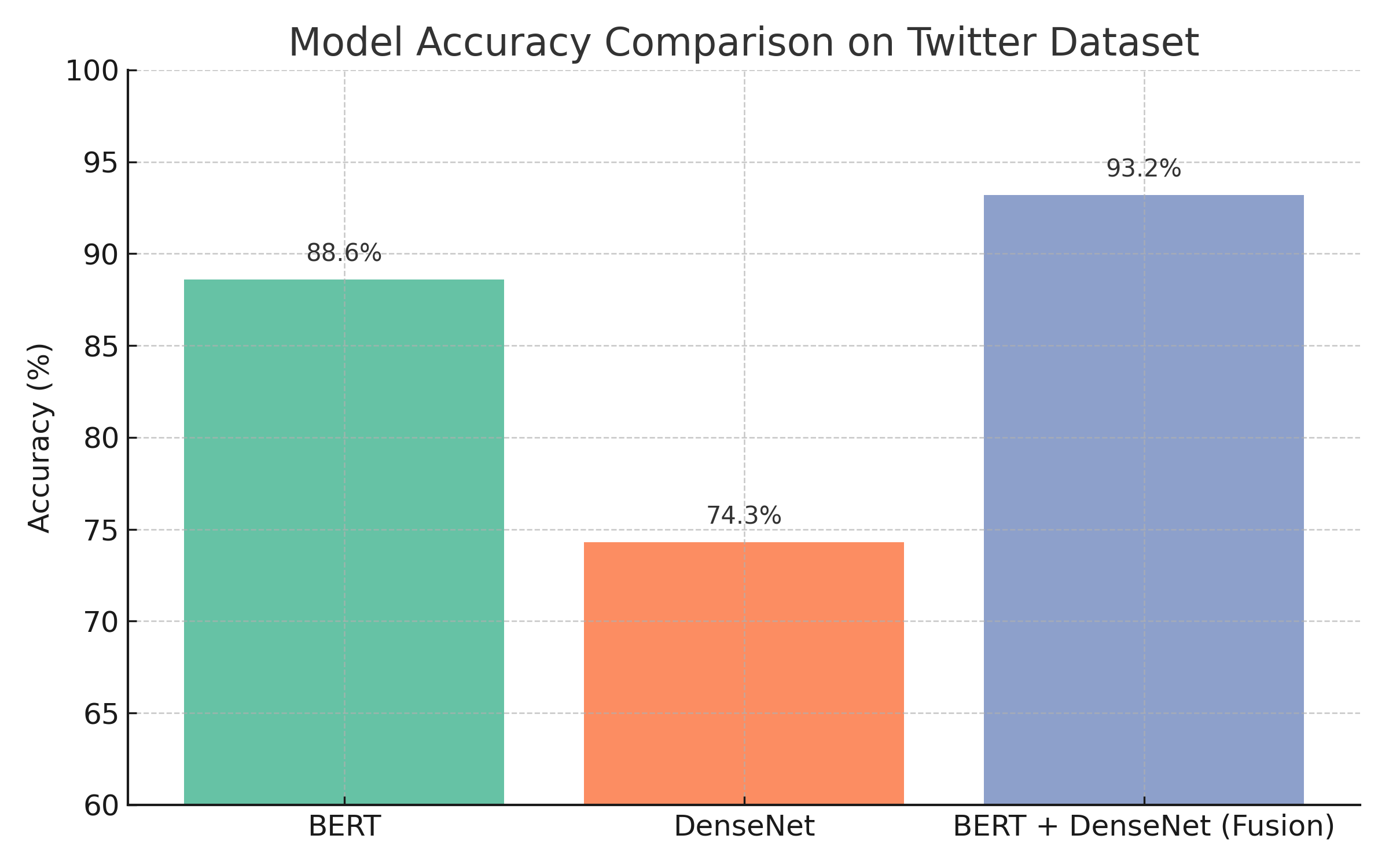}}
\caption{Model Accuracy Comparison on Twitter Dataset}
\label{fig:comparison}
\end{figure}

The results demonstrate clear advantages of multimodal learning. BERT alone achieved an accuracy of 88.6\%, DenseNet alone achieved 74.3\%, and the combined model reached 93.2\%. These findings confirm that visual signals add significant value in identifying sarcasm, especially when textual cues are ambiguous \cite{du2022effective}.

In a practical application scenario, this model could assist content moderators by automatically flagging sarcastic content, thus enhancing sentiment analysis systems. Future studies could expand this work by exploring multilingual sarcastic tweets, using emotion-labeled image datasets, and integrating audio cues such as tone and inflection \cite{zhang2019personalized}.

User-centered evaluation through crowdsourcing platforms (e.g., Amazon Mechanical Turk) can further validate the model’s predictions by comparing them with human judgments. This process enhances the credibility of sarcasm classification in real-world deployments \cite{calvo2017emotions}.

\section{Conclusion and Future Work}
We propose a conceptual sarcasm detection framework integrating sentiment, emotion, context, and linguistic cues through deep learning. The modular nature of the architecture enables flexibility in upgrading or replacing individual detection modules, making the system highly extensible and suitable for varied use cases. The integration of multiple feature domains ensures a holistic understanding of sarcasm, improving interpretability and robustness \cite{razali2021contextual}.

This research highlights the potential of combining advanced NLP techniques with deep learning models to enhance automated language understanding, particularly in nuanced areas like sarcasm. Our approach leverages pretrained models like BERT for deep contextual embeddings, complemented by handcrafted linguistic rules and emotion analysis \cite{liu2019multi, eke2021context, sharma2022hybrid}. The preliminary design showcases promising capabilities and adaptability across domains.

Future work involves implementing the full pipeline and conducting large-scale experiments across multiple datasets, including multilingual corpora \cite{barbieri2020tweeteval}. Real-time testing scenarios, such as integration with chatbots, virtual assistants, or sentiment analysis systems, are intended to validate the model's practical effectiveness \cite{chen2021chatbots}. We also aim to introduce adversarial training to make the model resilient against input manipulations and sarcasm obfuscation techniques \cite{du2022effective}.

Further improvements include enhancing multimodal detection by incorporating audio and video inputs. Prosodic features like tone, pitch, and speech rate can provide additional cues for sarcasm detection in voice-based systems. Likewise, visual elements like facial expressions and gestures could enrich interpretation in video communications \cite{zadeh2016multimodal}.

To address ethical considerations, future iterations will focus on fairness audits, bias mitigation, and explainability features \cite{calvo2017emotions}. Building user trust and ensuring transparent decision-making processes will be crucial, particularly in deployment scenarios involving moderation or user profiling. This research lays the groundwork for intelligent systems that can engage in human-like communication while maintaining ethical integrity and operational reliability.

\end{document}